\documentclass[runningheads]{llncs}
\usepackage[T1]{fontenc}
\usepackage{graphicx}
\begin{document}
\title{The Relationship Between Speech Features Changes When You Get Depressed: Feature Correlations for Improving Speed and Performance of Depression Detection}
%
%
\author{Fuxiang Tao\inst{1} \and
Wei Ma\inst{1} \and
Xuri Ge\inst{1}
\and
Anna Esposito\inst{2}
\and
Alessandro Vinciarelli\inst{1}}
\authorrunning{F. Tao et al.}
%

\institute{University of Glasgow, Glasgow G12 8QQ, UK 
\email{\{f.tao.1,W.ma.1,x.ge.2\}@research.gla.ac.uk}, \email{Alessandro.Vinciarelli@glasgow.ac.uk} \\ \and 
Universit\`a degli Studi della Campania ``Luigi Vanvitelli'', Caserta, Italy
\email{Anna.Esposito@unicampania.it}}
\maketitle              
\begin{abstract}
This work shows that depression changes the correlation between features extracted from speech. Furthermore, it shows that using such an insight can improve the training speed and performance of depression detectors based on SVMs and LSTMs.
The experiments were performed over the Androids Corpus, a publicly available dataset involving 112 speakers, including 58 people diagnosed with depression by professional psychiatrists. The results show that the models used in the experiments improve in terms of training speed and performance when fed with feature correlation matrices rather than with feature vectors. The relative reduction of the error rate ranges between 23.1\% and 26.6\% depending on the model. The probable explanation is that feature correlation matrices appear to be more variable in the case of depressed speakers. Correspondingly, such a phenomenon  can be thought of as a depression marker.

\end{abstract}
\section{Introduction}\label{intro}
The World Health Organization recently stated that depression affects 4.4\% of the world's population and accounts for 7.5\% of all years lived with disability~\cite{world2017depression}. The recent COVID-19 pandemics further aggravated such a situation and led to 53.2 millions more depression patients, an increase by 27.6\%~\cite{santomauro2021global}. Nevertheless, 63.6\% of the cases remain undiagnosed partly due to limited availability of healthcare services and partly because many patients lack financial resources necessary to access appropriate medical attention~\cite{faisal2022depression}. For these reasons, the computing community made major efforts towards the development of depression detection technologies.

Many approaches use speech as input and the reason is that the pathology was shown to leave traces in the way people speak (see Section~\ref{Work}).
In fact, according to neuroscience, brain connectivity patterns tend to be more unstable in depression patients~\cite{wise2017instability}. 
In particular, depressed speakers tend to show a lower degree of coordination across different brain areas and this can ``[...] alter speech production by influencing the characteristics of the vocal source, tract, and prosodics [and lead to] psychomotor retardation, where a patient shows sluggishness and motor disorder in vocal articulation, affecting coordination across multiple aspects of production''~\cite{williamson2013vocal}. Such a phenomenon is expected to change not only acoustic properties of speech, but also the relationship between them. Correspondingly, it is expected to change the correlation between features and the way such a correlation changes over time. 


Despite the above, to the best of our knowledge, the literature pays only limited attention to the relationship between features (see Section~\ref{Work}). The key assumption of this work is that this is an issue because it results into missing the major depression effects described earlier.
For this reason, the focus of this work is on investigating whether feature correlations and their changes over time can help improving the effectiveness of depression detection. 

The experiments were performed over the Androids Corpus, a publicly available dataset involving 112 speakers. The focus is on the read speech samples for two main reasons.
The first is that the brain phenomena mentioned earlier were observed also when people read~\cite{regev2013selective}, the second is that read speech involves less variability resulting from sources not necessarily related to the pathology (e.g., the topic being discussed). The speech samples were represented as sequences of feature correlation matrices (see Section~\ref{approach}) and such a representation was shown to improve two classification approaches, namely Support Vector Machines (SVM) and Long Short-Term Memory networks (LSTMs) ~\cite{hochreiter1997long}. Compared to feature vectors, correlation matrices were shown to reduce the error rate by up to 26.6\% and up to 23.1\% for SVMs and LSTMs, respectively. Overall, the proposed approaches reached an F1 score of up to 87.7\% when fed with feature correlation matrices and up to 83.7\% when fed with feature vectors. In addition, using sequences of feature correlation matrices made the classification process roughly two times faster.

Overall, the main contributions of this work can be summarized as follows:
\begin{itemize}
\item To the best of our knowledge, this is the first work comparing the performance of the same models (SVM and LSTM) when being fed with feature vectors and with feature correlation matrices;
\item This is the first work showing that changes in correlation patterns over time can act as a depression marker.
\end{itemize}
The comparison with previous results obtained over the same data shows that the novelties above lead, on average, to higher performances.

The rest of this article is organized as follows: Section~\ref{Work} describes related work, Section~\ref{data} describes the data, Section~\ref{approach} describes the approach used in the experiments, Section~\ref{expres} reports on experiments and results, and the final Section~\ref{concl} draws some conclusions.
\section{Related Work}\label{Work}
In recent years, the computing community made substantial efforts towards automatic depression detection and identification of depression markers in speech~\cite{Cummins2015,wu2022automatic,highland2022review}. 
After observing that speech production is different in depression patients~\cite{Cummins2015}, several works tried to identify markers related to temporal properties. For example, it was shown that the effectiveness of depression detection can be improved significantly by taking into account speech rate~\cite{morales2016speech} or frequency of pauses~\cite{tao2020spotting}. However, this type of marker was shown to have different effectiveness in correspondence of different types of speech~\cite{kiss2017comparison}.

In most cases, the approaches do not try to identify markers, but  represent speech in terms of feature vectors and then use different machine learning methodologies to identify depression patients. The most common features are, by far, Mel Frequency Cepstral Coefficients (MFCCs), often shown to be beneficial for the performance~\cite{Cummins2015,wu2022automatic,taguchi2018major,rejaibi2022mfcc,tao2023multi}. MFCCs, and any other Low-Level Descriptors, are typically extracted from short analysis windows and they are often concatenated with their derivatives to account for possible temporal effects. Such an approach was shown to be effective for both spontaneous and read speech~\cite{Cummins2015,cummins2011investigation,alghowinem2013detecting}.

Correlation between features was explored, to a limited extent, as a possible way to capture ``[...] psychomotor retardation, where a patient shows sluggishness and motor disorder in vocal articulation, affecting coordination across multiple aspects of production''~\cite{williamson2013vocal}. The proposed approach was based on measuring the correlation between features extracted at multiple time distances, thus resulting in multiple correlation matrices concatenated with each other~\cite{williamson2013vocal}. The approach was shown to be effective not only for depression, but also for epileptic seizures~\cite{williamson2011epileptic}. More recent studies show that such a correlation representation is effective with deep neural networks too~\cite{song2020spectral,huang2020exploiting,ge2019colloquial}. The main issues are the need for longer speech recordings (the time distances for measuring the correlation can be high)~\cite{williamson2016detecting} and the appearance of artefacts that require the application of filters~\cite{huang2019investigation}. The approach proposed in this work addresses both problems by measuring the correlation over short intervals of time and by modelling the correlation changes through sequential models. 
\section{The Data}\label{data}
\begin{table}[t!]
\begin{center}
\caption{The table provides demographic information about the participants. Low and High refer to the education level (1 participant did not disclose information about their studies and, therefore, the sum for columns Low and High is 111 rather than 112).}\label{tab:participant}
\begin{tabular}{cccccc}
\hline
 & Age & Male & Female & Low & High \\
 \hline
Control & 47.1 $\pm$ 12.8 & 12 & 42 & 19 & 35 \\
Depression & 47.4 $\pm$ 11.9 & 20 & 38 & 25 & 32 \\
Total & 47.2 $\pm$ 12.3 & 32 & 80 & 44 & 67 \\
\hline
\end{tabular}
\end{center}
\end{table}
The experiments were performed over the read speech samples of the Androids Corpus, a publicly available dataset~\cite{tao2023corpus}. Overall, the corpus involves 118 participants, but the experiments focus on the 112 who provided read speech samples.
Each participant is a native Italian speaker and was asked to read aloud an Aesop's tale in Italian (``\emph{The North Wind and the Sun}''). Informed consent was obtained from all participants before the experiment, in accord with privacy and data protection laws in Italy, the country where the data was collected. The reason for choosing the fairy tale above is that it is easy to understand and it does not contain words that an average reader might not know (the version of the tale used in the experiments was extracted from a book for children). This ensures that the participants can understand the text irrespective of their education level. In order to simulate the normal setting in which depression patients interact with doctors, the recordings were collected  with a standard laptop microphone in the clinical consultation rooms of the three Mental Health Centers involved in the study.

Out of the 112 participants, 58 reported no history of disorders (neurological or psychiatric) and made no use of medications or recreational drugs (referred to as \emph{control} participants hereafter). The remaining 54 were diagnosed with depression by professional psychiatrists using the \emph{Diagnostic and Statistical Manual of Mental Disorders 5} (DSM-5).
Table~\ref{tab:participant} shows information about age, gender and education level. Statistical analysis shows that there is no difference in age distribution between depressed and control participants ($p > 0.05$ according to a two-tailed $t$-test). Similarly, there is no statistically significant difference in terms of gender and education level according to $\chi^2$ tests. This suggests that speech differences between the two groups of participants depend on depression and not on other factors. 

The age distribution of the data matches the age range of people that tend to develop depression more frequently~\cite{aloshban2021you}. Similarly, the number of female depressed participants is close to 2 times greater than male ones. This is in line with epidemiological observations showing that depression is more common among women than among men~\cite{kessler2003epidemiology}. In this respect, the sample is expected to represent the general population of both depressed and non-depressed individuals. The total duration of the recordings is 1 hour, 33 minutes and 49 seconds, with an overall average of 50.3 seconds. When considering separately control and depressed participants, the averages are 52.9 and 47.4 seconds, respectively (the difference is statistically significant with $p < 0.01$ according to a two-tailed $t$ test). 
\section{The Approach}\label{approach}
\begin{figure}[t]
  \centering
  \includegraphics[width=0.95\linewidth]{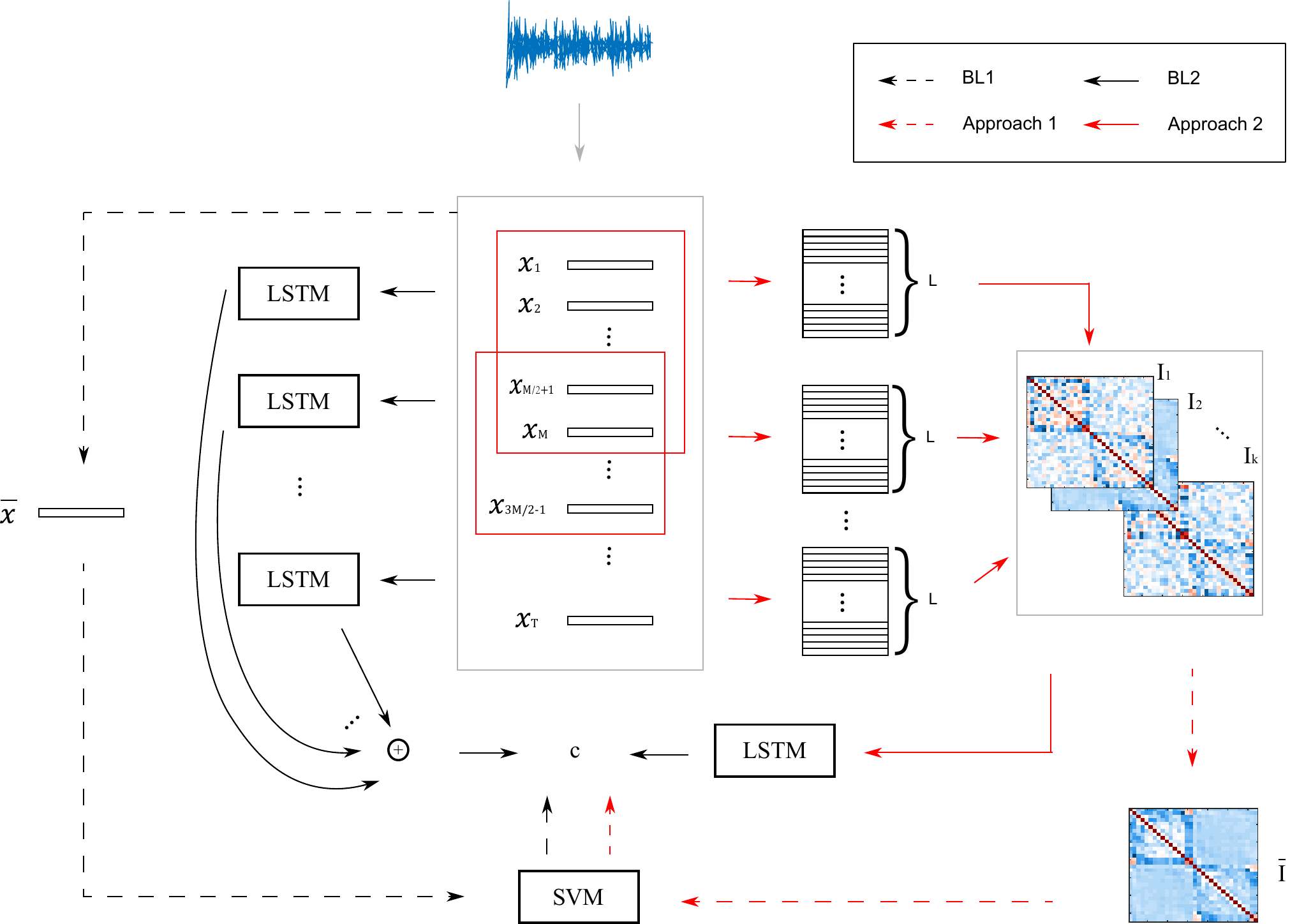}
  \caption{The diagram shows the four approaches used in the experiments. The black arrows stand for the use of feature vectors while the red arrows stand for the use of correlation matrices. The black dashed arrows stand for BL1, while the black arrows stand for BL2. The red dashed arrows stand for Approach 1, while the red arrows stand for Approach 2. The symbol $\oplus$ corresponds to aggregation.$c$ is the class assigned by the classification.}\label{diagram}
\end{figure}

The goal of the experiments is to test whether feature correlation matrices convey more depression-relevant information than feature vectors. For this reason, the approach (see Figure~\ref{diagram}) includes a \emph{feature extraction} step (conversion of speech signals into sequences of feature vectors), a \emph{correlation representation} step (mapping of sequences of feature vectors into sequences of feature correlation matrices) and a \emph{depression detection} step. Figure~\ref{diagram} shows that this latter can be performed by feeding the same model (SVM or LSTM) with either feature vectors or correlation matrices. In this way, it is possible to test whether these latter convey more information than the sole features.

The aim of the feature extraction is to convert the speech recordings into sequences of feature vectors $\vec x_k$. In the experiments of this work, the vectors were extracted from $25$ ms long analysis windows at regular time steps of $10$ ms. Both values are standard in the literature and no attempts were made to identify alternatives possibly leading to better results. The extraction was implemented with OpenSMILE~\cite{eyben2013recent} 
and made use of a $32$-dimensional feature set widely applied in emotion recognition~\cite{schuller2009interspeech}. The $32$ features include the following:
\begin{itemize}
\item Root Mean Square of the Energy (Energy): related to loudness and it tends to be lower for depressed individuals~\cite{schuller2013computational};
\item Mel-Frequency Cepstral coefficients 1-12 (MFCC): they account for the phonetic content of the data and they are widely used in depression detection~\cite{Cummins2015}; 
\item Fundamental Frequency (F0): frequency that carries the highest energy in the signal~\cite{Quatieri2012};
\item Zero-Crossing Rate (ZCR): it indicates how many times a signal crosses the value of zero per millisecond and accounts for F0~\cite{Nilsonne1985};
\item Voicing probability (VP): it accounts for the probability of a frame corresponding to emission of voice and it was shown to account for pauses that can help one to discriminate between depressed and non-depressed speakers~\cite{tao2020spotting};
\end{itemize}
The 16 features are doubled by taking into account the differences between the feature values extracted from two consecutive analysis windows, thus leading to vectors of dimension $D=32$.

The correlation representation step segments the sequence of feature vectors into subsequences of length $L$ (the number of vectors included in one subsequence) starting at regular steps of length $L/2$ (two consecutive subsequences overlap by half of their elements). In the experiments, the value of $L$ ranges between $100$ and $500$, corresponding to time intervals of length between $1$ and $5$ seconds. Once the segmentation is performed, it is possible to extract a \emph{local feature correlation matrix} $I_k$ in which element $\{I_k\}_{ij}$ is the correlation between extracted features $i$ and $j$ ($i,j \in [1,D]$) along the subsequence. Given that there are multiple subsequences, there will be multiple local feature correlation matrices $I_k$, with $k$ ranging between $1$ and $T$, where $T$ is the total number of subsequences. Matrix $I_n$ corresponds to the subsequence starting at vector $(n-1)L/2$ and ending at vector $[(n-1)/2+1]L -1$, where $n=1,\ldots,T$. All correlation coefficients are converted into $Z$-scores with Fisher’s transformation, a standard step in statistical analysis of dependent correlations~\cite{meng1992comparing}.

A first baseline approach for depression detection is (BL1) in~\cite{tao2023corpus} that takes the average of the feature vectors extracted from a recording and feeds it to a linear kernel SVM implemented with scikit-learn (version in 0.23.2)~\cite{pedregosa2011scikit}, the BL1 in Figure~\ref{diagram}. Similarly, given that the correlation representation step converts every speech signal into a sequence of matrices $I_k$, it is possible to estimate the average $I_k$ and feed it to a Support Vector Machine to perform the depression detection step (see Approach 1 in Figure~\ref{diagram}). The correlation matrices are symmetric and only the elements below the principal diagonal are used, thus leading to a dimension $D(D-1)/2 = 496$ ($D=32$ is the original number of features). BL1 and Approach 1 can be compared to test whether feature correlation matrices are actually of help.

Another way to perform depression detection is to split the sequence of the feature vectors into subsequences of length $128$ (a standard value in the literature) and then feed each one of these to LSTMs (BL2 in~\cite{tao2023corpus}) implemented with PyTorch (version in 1.13.1+cu116)~\cite{paszke2019pytorch}. Once all subsequences are classified, it is possible to apply a majority vote and assign a recording to the class its subsequences are more frequently assigned to (BL2 in Figure~\ref{diagram}). The same LSTMs can be applied to the sequence of the $I_k$. In this case, a linear layer reducing the dimension of the correlation matrices to $32$ is added to the LSTMs with the goal of making the problem computationally more tractable (Approach 2 in Figure~\ref{diagram}). BL2 and Approach 2 can be compared to test whether the matrices actually improve over the feature vectors. 
\section{Experiments and Results}\label{expres}
\begin{table}[t!]
\begin{center}
\renewcommand\tabcolsep{3.0pt}
\caption{Depression detection results in terms of Accuracy, Precision, Recall and F1 score. The baselines and approaches are numbered 1 to 2 according to Figure~\ref{diagram}. The performance metrics are represented in terms of average and standard deviation obtained over 10 repetitions of the experiment. At every repetition, the weights of the LSTM were initialized to different random values. The table reports the best accuracy over different lengths $L$.}\label{tab:accuracy}
 \begin{tabular}{ccccc}
 \hline
 & Accuracy & Precision & Recall & F1 Score \\ 
 \hline
Random & 50.1 & 51.8 & 51.8 & 51.8 \\
BL1 & 69.6$\pm$5.3 & 73.6$\pm$19.1 & 68.8$\pm$12.0  & 68.4$\pm$7.7 \\
Approach 1 & 77.7$\pm$5.4 & 78.1$\pm$14.2 & 83.1$\pm$13.0  & 78.5$\pm$4.9 \\ 
BL2 & 84.4$\pm$1.1 & 84.5$\pm$2.1 & 85.6$\pm$2.8  & 83.7$\pm$1.1 \\ 
Approach 2 & 88.0$\pm$1.0 & 88.1$\pm$1.1 & 89.2$\pm$1.9  & 87.7$\pm$1.7 \\
\hline
\end{tabular}
\end{center}
\end{table}

The experiments were performed according to the $k$-fold protocol ($k=5$) available in the Androids Corpus distribution. 
The folds are disjoint so that 
the same participant never appeared in both training and test set. This ensures that the approach actually detects depression and does not simply recognize the speaker. 

The number of hidden states in the LSTMs was set to $H = 32$ (the same number of hidden states as the baselines in the Androids Corpus~\cite{tao2023corpus}). The number of training epochs was $T=100$ and the learning rate was set to $0.0005$. The training was performed using the RMSProp as an optimizer~\cite{tieleman2012lecture} and the categorical cross-entropy as a loss function. All experiments were replicated $R=10$ times using a different initialization of the networks. For this reason, all results are reported in terms of average and standard deviation observed over the repetitions. This ensures that the performances are not the result of a favorable initialization, but a realistic estimate of the system's effectiveness.

Table \ref{tab:accuracy} shows the results in terms of Accuracy, Precision, Recall and F1 Score. The random baseline approach assigns an unseen sample to a certain class with probability corresponding to its prior. This leads to the following accuracy:
\begin{equation}
\hat\alpha=p(c)^2+p(d)^2, 
\end{equation}
where $p(c)$ is the prior of class \emph{control} and $p(d)$ is the prior of class \emph{depressed}. The corresponding Precision, Recall and F1 score are all equal to $p(d)$. 
The table shows the performance of SVMs and LSTMs when giving as input both the sequence of vectors $\vec x_k$ and the sequence of feature correlation matrices $I_k$. According to a two-tailed $t$-test, all approaches perform better than the random baseline ($p<0.001$ in all cases). Furthermore, always according to a two-tailed $t$-test, models perform better when fed with feature correlation matrices than when fed with feature vectors. The error rate decreases by 26.6\% when passing from BL1 to Approach 1. Similarly, the relative reduction of the error rate is up to 23.1\% for approaches based on LSTMs. In this respect, the results suggest that feature correlation matrices actually convey more depression-relevant information than simple feature vectors. 
\begin{figure}[t]
\centering
\includegraphics[width= 0.99\linewidth]{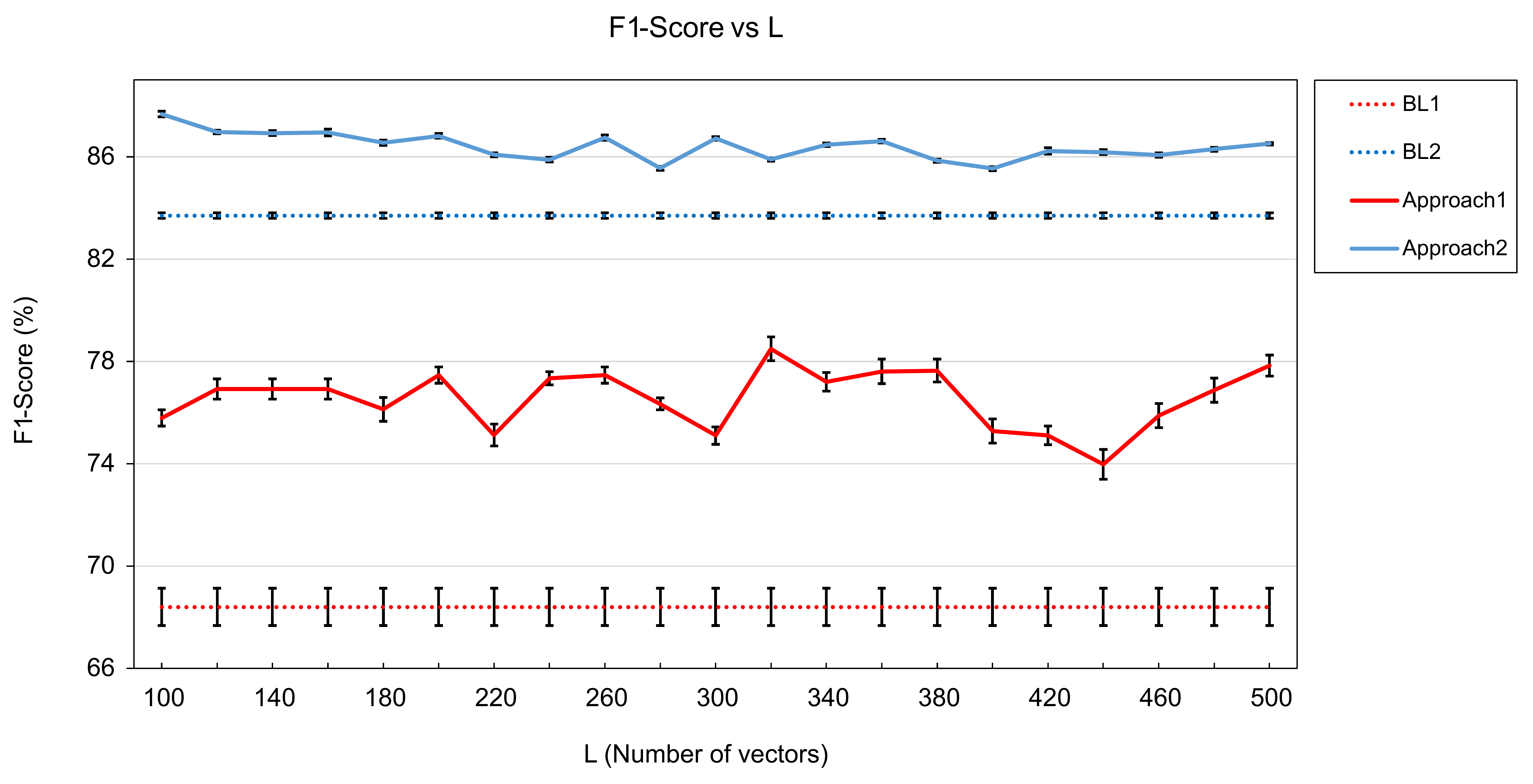}
\caption{The figure shows averaged F1-Score and its standard error of the mean over 10 repetitions at different lengths $L$.}\label{fig:recognition_plot}
\end{figure}

Given that BL1 and Approach 1 do not include the step of converting vectors to matrices, the results of these approaches do not change with parameter $L$. Table~\ref{tab:accuracy} shows the best results across all values of parameter $L$. For this reason, Figure~\ref{fig:recognition_plot} shows how the F1-Score changes depending on $L$ and the results suggest that the models fed with feature correlation matrices tend to outperform those fed with feature vectors, thus confirming that the results of Table~\ref{tab:accuracy} are not the result of a particular choice of $L$. 

In addition, the use of the feature correlation matrices reduces significantly the amount of time required to train the LSTMs. When using the same computing infrastructure (Google CoLab Tesla T4 GPU), the time goes from 4.5 to 2 minutes for training LSTMs, such an improvement is statistically significant according to a two-tailed $t$-test ($p < 0.001$). This is mainly because LSTMs need less cells in the case of correlation matrix sequences. In the case of SVMs, no significant difference was observed in terms of training time. A further advantage of correlation matrix sequences is that, being shorter, they do not need to be segmented into subsequences to be fed to the LSTMs. Therefore, it is possible to avoid the majority vote.

\begin{figure*}[t!]
  \centering
  \includegraphics[width=0.99\linewidth]{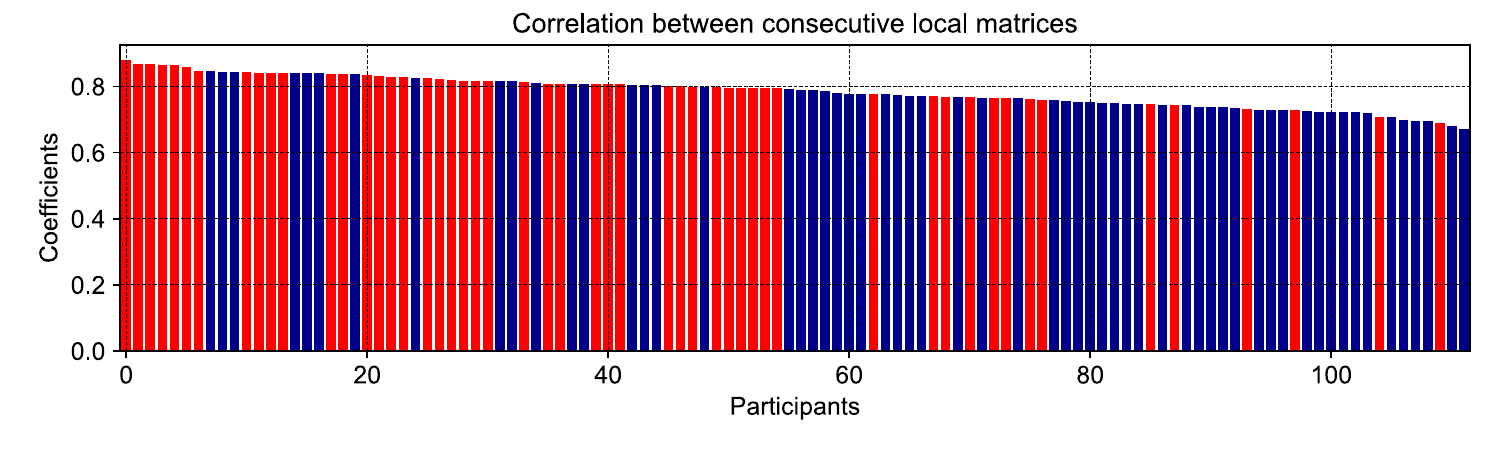}
  \caption{The chart shows the average correlation between consecutive feature correlation matrices for control (red bars) and depressed (blue bars) participants, ordered from highest to lowest. Control speakers tend to be more frequent in the first part of the chart and this suggests that the correlation tends to be higher for them.}\label{matrix}
\end{figure*}

One possible explanation of the results above is in Figure~\ref{matrix}. The chart shows the average Spearman correlation coefficients between consecutive correlation matrices $I_k$ and $I_{k+1}$ (the elements below the principal diagonal). A two-tailed $t$-test shows that, for all values of $L$, the average correlation between consecutive correlation matrices is higher, to a statistically significant extent, in the case of control participants ($p<0.05$ with FDR correction for all values of $L$). In particular, the correlation is around $0.75$ for control participants and around $0.71$ for depressed ones, for all values of $L$. Such an observation suggests that the relationship between features extracted at different points in time tends to be less consistent in the case of depressed people and this is probably a depression marker that helps the models to perform better.
\section{Conclusions}\label{concl}
The key-result of this paper is that feature correlation matrices lead to better depression results than feature vectors, at least in the case of linear kernel SVMs and LSTMs, the two models used in the experiments. Furthermore, the proposed end-to-end approach significantly reduces the time for training. To the best of our knowledge, this is the first work that compares feature vectors and correlation matrices in terms of the performance they lead to. Furthermore, this is the first work proposing an explanation of the observed results in  terms of a possible marker (the different correlation between consecutive matrices).

The main limitation of the experiments is the use of a linear layer in Approach 2. Its aim is to keep the same input dimensionality as BL2, while still preserving the correlation between the features. This costs extra-parameters that make it less clear, in the comparison between BL2 and Approach 2, whether the performance improvement actually results from  the correlation matrices. On the other hand, in the case of the SVMs, the only change between BL1 and Approach 1 is the use of the matrices and the improvement is statistically significant. This seems to confirm that the matrices actually help to improve depression detection.




%
\begin{table}[t!]
\begin{center}
\caption{Previous studies involving the same participants. 
}\label{tab:stateofart}
\begin{tabular}{ccccc}
\hline
Studies                                                 & Accuracy & Precision & Recall & F1 \    \\ \hline
Scibelli et al.,~\cite{scibelli2018depression}  & 77.0     & 74.0      & 80.0   & 77.0 \\
Tao et al.,~\cite{tao2020spotting}         & 84.5     & 84.5      & 84.6   & 84.5 \\
Aloshban et al.,~\cite{aloshban2020detecting}  & 83.5     & 95.0      & 70.3   & 80.5 \\
Aloshban et al.,~\cite{aloshban2021language}   & 83.0     & 95.2      & 69.0   & 80.0 \\
Aloshban et al.,~\cite{aloshban2021you}        & 84.7     & \textbf{95.4}      & 72.4   & 82.3 \\
Alsarrani et al.,~\cite{alsarrani2022thin}       & 67.6 & 71.7 & 72.2 & 72.3 \\
Tao et al.,~\cite{tao2023corpus}         & 84.4 & 84.5 & 85.6  & 83.7 \\
Ours  & \textbf{88.0} & 88.1 & \textbf{89.2} & \textbf{87.7} \\ \hline
\end{tabular}
\end{center}
\end{table} 

Table~\ref{tab:stateofart} shows results obtained in other works using the data of the Androids Corpus. A fully rigorous comparison with previous approaches is not possible because the experimental protocol is not always the same (see below for more details), but the performances presented in this work are the highest in terms of Accuracy, Recall and F1 Scores, thus confirming that correlation matrices are of help. In two cases~\cite{scibelli2018depression,tao2020spotting}, the data is the same (read speech), but not all the speakers analyzed in this work were involved. In~\cite{alsarrani2022thin}, the speakers are the same, but the data is different (spontaneous speech). Finally, the results in~\cite{aloshban2020detecting,aloshban2021language,aloshban2021you} were obtained over 59 speakers only and using not only paralanguage, but also language. The only rigorous comparison is with the results in~\cite{tao2023corpus} (same data, speakers and experimental protocol).


This work focused on read speech and, therefore, one possible direction for future work is the application of correlation matrices to spontaneous speech data. The aim of this work is to examine the effectiveness of correlation matrices by comparing them with feature vectors, but it never involved  the combination of both. Correspondingly, another possible future direction is to use both features and their correlation matrices to detect depressed speakers.

\subsubsection{Acknowledgements} The research leading to these results has received funding from the project ANDROIDS funded by the program V:ALERE 2019 Universit\`a della Campania “Luigi Vanvitelli”,  D.R. 906 del 4/10/2019, prot. n. 157264,17/10/2019. The work of Alessandro Vinciarelli was supported by UKRI and EPSRC through grants EP/S02266X/1 and EP/N035305/1, respectively.

%
%
%
\bibliographystyle{splncs04}
\bibliography{mybib}
%





\end{document}